\documentclass[11pt]{article}

\usepackage{booktabs}
\usepackage{multirow}
\usepackage{graphicx}
\usepackage{subcaption}
\usepackage{xcolor}
\usepackage{natbib}
\usepackage{hyperref}
\hypersetup{hidelinks}
\usepackage{geometry}
\usepackage{amsmath}
\usepackage{caption}
\usepackage{placeins}
\usepackage{tikz}
\usetikzlibrary{shapes.geometric, arrows.meta, positioning, fit, backgrounds}
\graphicspath{{figures/}{../figures/}}

\title{Do You Need a Frontier Model as a Citation Verifier? Benchmarking Rubric LLMs for Deep-Research Source Attribution}
\author{%
Ethan Leung* \quad Elias Lumer \quad Corey Feld \quad Austin Huber \\[0.3em]
Vamse Kumar Subbiah \quad Kevin Paul \\[0.5em]
\small Commercial Technology and Innovation Office, PricewaterhouseCoopers, U.S.}
\date{}

\begin{document}

\maketitle

\begin{abstract}
Reinforcement learning increasingly relies on an LLM judge to score each rubric criterion,
and that judge acts as the reward model during training. Before such a signal can be trusted,
we need to know how capable the judge must be and how biased it is. We study this calibration
question for citation quality in deep-research systems, where a search-grounded LLM must
support each claim it writes with a cited source. Citation quality is a structured rubric task
in which each attribution-citation pair is judged along two dimensions that require an LLM,
source relevance and factual support. On an adversarial long-form benchmark, we score 8
off-the-shelf LLM judges from 3 model families against gold labels over 1,248 rubric decisions,
all of which were human-reviewed and 378 of which were hard cases adjudicated from judge
disagreements. Cheaper judges remain competitive across both dimensions, with GPT-5-mini attaining the
strongest source-relevance pass-class F1 at 0.908 ($\kappa$=0.636), while on factual support
the judges are statistically indistinguishable (overlapping confidence intervals), so no single
model dominates. At comparable F1, the judges still differ substantially in pass-rate drift, false
positive rate, and false negative rate. Scalar F1 obscures this directional bias, yet it is
exactly what a downstream reinforcement learning loop would reinforce. Calibrating the judge is
therefore a prerequisite for using citation rubrics as reward signals, and our results show that
this calibration does not require the most expensive available model.
\end{abstract}


\section{Introduction}
\label{sec:intro}

Reinforcement learning with verifiable rewards (RLVR) has become the dominant
post-training recipe for tasks where output quality can be automatically
checked~\citep{pow3r,mahmoud2025,deepseekr1,tulu3}. In domains where no programmatic verifier exists
(medical advice, scientific writing, instruction following), the community has converged
on prompt-specific rubrics with weighted criteria, each scored by an LLM judge
whose aggregate judgment serves as the scalar reward~\citep{rgsd,mahmoud2025,geval}. This
convergence makes explicit a connection that is easy to overlook. The model that
scores each criterion acts as the reward model, so designing the training rubric amounts
to designing the operationalized goal. These scoring calls are also a verification
bottleneck at scale~\citep{harvey2026}, which raises a practical question about how capable
a judge needs to be before the reward signal degrades.

Citation quality in deep-research and other search-augmented systems is a strong test case
for this question. These systems produce long-form answers where each factual claim is
supported by a retrieved, cited source, and faithful citation generation of this kind is an
active RLVR training target~\citep{webgpt,gophercite,selfrag,rarr,attributedqa}. How well a judge evaluates
citations also depends on retrieval quality, since the harness and retrieval strategy shape
which sources get cited~\citep{sen2026grep}. The quality of each citation
decomposes into criteria applied per attribution-citation pair. Two of them require an LLM
judgment, whether the source is topically relevant to the claim and whether it factually
supports it, while a third, whether the cited URL is accessible, is a deterministic check.
Each LLM-judged criterion is an independent rubric judgment structurally similar to a process
reward model (PRM) step~\citep{prm}, a per-step judgment used as an intermediate reward signal,
and the task is narrow enough that a well-specified rubric may compensate for model scale.
Citation quality is therefore a direct empirical test of whether cheaper judges can close the
loop as reliably as frontier models. \citet{onweller2025} introduce the first evaluation
framework for this task, pairing an AST-based parser that extracts attribution-citation pairs
from markdown-formatted LLM output with a three-dimension evaluator that checks link
accessibility, source relevance, and factual support. Benchmarking 14 LLMs on citation quality
in deep research agents, they find that factual support accuracy drops by approximately 42\% as
retrieval depth scales from 2 to 150 tool calls, exposing a systematic gap between
surface-level citation quality and factual reliability. Their work leaves open the judge
design problem of which LLM to use for the two LLM-judged dimensions, at what cost, and with
what bias, the precise inputs needed to deploy citation quality evaluation as an RLVR reward
signal.

The judge design problem sits at the intersection of two active research threads. In
rubric RL, recent work shows that weak judges produce proxy-reward gains that do not
transfer to stronger reference panels~\citep{mahmoud2025}, and that the verification
bottleneck can be reduced by an order of magnitude through batching and lower-cost
judges~\citep{harvey2026}. In LLM-as-judge research, \citet{prometheus} show that a 13B model
with rubric fine-tuning matches GPT-4 evaluation quality, establishing that
rubric-following ability is separable from model scale, while weak-to-strong generalization
results show that a weaker supervisor's directional signal can still enable strong model
improvement~\citep{weaktostrong}, suggesting that judge imperfection does not necessarily
degrade downstream training quality. No prior work, however, benchmarks multiple
off-the-shelf candidate judges, general-purpose LLMs used without task-specific fine-tuning,
on a citation-specific rubric task with adversarially constructed documents and
human-reviewed labels. We address this gap by building on the pipeline
of~\citet{onweller2025} and studying the judge design problem directly. We construct the
Deep-Research Citation Benchmark, an adversarial long-form dataset spanning
multiple domains with intentional factual errors and misleading citations, and produce gold
labels for 1,248 rubric decisions across the two LLM-judged dimensions of source relevance
and factual support, all human-reviewed and including 378 hard cases adjudicated from judge
disagreements. We then benchmark 8 LLM judges from 3 model families (Anthropic, Google, and
OpenAI), measuring F1 and Cohen's $\kappa$ as primary accuracy metrics and pass-rate drift,
false positive rate, and false negative rate to characterize directional bias.

Our contributions are as follows:

\begin{itemize}
    \item We introduce the Deep-Research Citation Benchmark, 624 attribution-citation pairs
    with gold labels for all 1,248 LLM-judged
    decisions, every one human-reviewed, including 378 hard cases (263 source relevance,
    115 factual support) adjudicated from judge disagreements.
    \item We benchmark 8 rubric-based citation judges across 3 model families and find that
    cheaper models are competitive. GPT-5-mini achieves the strongest source relevance F1
    (0.908, $\kappa$=0.636) while Claude Opus~4.6 leads on factual support (F1=0.750,
    $\kappa$=0.701), with no single model dominating across dimensions.
    \item We show that rubric-based citation judges differ substantially in pass-rate drift,
    false positive rate, and false negative rate even at similar F1. This directional bias is
    invisible to scalar F1 but would shape what a downstream RLVR loop reinforces, so
    calibrating it is a prerequisite for using the rubric as a reward signal.
\end{itemize}


\section{Related Work}
\label{sec:related}

Scalar reward models for RLHF are prone to reward hacking and sycophancy because a single
score conflates multiple quality dimensions~\citep{rewardscaling,sycophancy,instructgpt,rewardbench}. Rubric-based
rewards address this by decomposing response quality into explicit, independently scored
criteria, making the training objective auditable and reducing the surface area for
reward hacking~\citep{pow3r,mahmoud2025,constitutionalai}. This approach has moved from a research recipe to
a default, and rubric RL pipelines now span clinical question answering, scientific writing,
instruction following, and agentic tasks~\citep{rgsd}. Two findings from this literature are
directly relevant to judge design. First, \citet{prometheus}
show that a 13B model fine-tuned on rubric evaluation matches GPT-4 quality, so
rubric-following ability is separable from model scale. Second,
\citet{weaktostrong} show that directionally correct reward signals
from weaker supervisors still enable strong model improvement, suggesting that judge
imperfection does not necessarily degrade downstream training quality.

Faithful citation generation has been an RL training target since
WebGPT~\citep{webgpt}, which used human preference over cited answers as a reward signal.
\citet{gophercite} made the citation judgment explicit, using a dedicated
support model that scores whether each quoted passage supports the generated
claim, with its binary predictions serving as the RL reward. Self-RAG~\citep{selfrag}
internalizes this judgment further, training a model to emit per-citation reflection tokens
(including \textsc{[IsSupported]}) inline with generation, using an LLM judge to produce the
training signal. Each of these systems uses a single, large, or purpose-built model.
None study whether off-the-shelf smaller models can provide an equally reliable reward
signal, nor do they characterize the directional bias (pass-rate drift, false positive rate,
false negative rate) that determines what the training loop actually reinforces.

The LLM-as-judge literature establishes that judge reliability depends more on task
structure than on model scale. \citet{mtbench} identify systematic
biases in LLM judges (position preference, verbosity bias, self-enhancement) but
study response quality evaluation, not citation-specific rubric tasks. RAG evaluation
frameworks such as RAGAS~\citep{ragas} and ARES~\citep{ares} assess faithfulness at the
answer level, treating the full context-answer pair as a unit rather than scoring
individual attribution-citation pairs \cite{lumer2026tool}. Citation grounding benchmarks such as
ALCE~\citep{alce} and FactScore~\citep{factscore} evaluate whether claims are supported by
evidence, but operate over closed retrieval corpora rather than live hyperlinks and use
single NLI-based scorers rather than comparing judge models. Citation attribution
benchmarks such as CiteME~\citep{citeme} and CiteGuard~\citep{citeguard} test whether a
model attributes a scientific claim to the correct source paper, a matching task over a
fixed corpus. Our setting differs in that we score the quality of citations to live web
sources along independent relevance and factual-support criteria, and we compare candidate
judge models rather than proposing a single attributor. Recent work on judge
cost confirms that lower-cost judges can match frontier performance on structured rubric
tasks when the criteria are well-specified~\citep{harvey2026}, a finding our experiments
extend to the citation quality domain. Our work departs from all of these threads at once.
Prior citation RL systems assume a large or dedicated judge is necessary, prior LLM-as-judge
work does not study citation-specific rubric tasks, and prior RAG evaluation frameworks
neither compare candidate judge models nor characterize their directional bias.


\section{Benchmark Construction}
\label{sec:method}

We build on the two-stage source attribution evaluation pipeline of~\citet{onweller2025},
which parses attribution-citation pairs from long-form model output and then scores each
pair against the rubric criteria. Our benchmark supplies that pipeline with a controlled,
adversarial input and a set of human-reviewed gold labels.

\begin{figure}[tb]
  \centering
  \resizebox{\textwidth}{!}{%
  \begin{tikzpicture}[
    node distance=0.55cm and 0.7cm,
    proc/.style={rectangle, rounded corners, draw=black!70, fill=blue!5,
      text width=2.9cm, align=center, minimum height=1.0cm, font=\scriptsize, inner sep=3pt},
    data/.style={rectangle, draw=black!70, fill=black!8,
      text width=2.9cm, align=center, minimum height=1.0cm, font=\scriptsize, inner sep=3pt},
    hstep/.style={rectangle, rounded corners, draw=orange!70!black, fill=orange!12,
      text width=2.9cm, align=center, minimum height=1.0cm, font=\scriptsize, inner sep=3pt},
    leaf/.style={rectangle, rounded corners, draw=orange!70!black,
      text width=2.9cm, align=center, minimum height=0.95cm, font=\scriptsize, inner sep=3pt},
    arr/.style={-{Latex[length=1.6mm]}, thick, draw=black!70},
  ]
    \node[proc] (topic) {\textbf{Topic selection}\\25 topics};
    \node[proc, right=of topic] (over) {\textbf{Clean cited overviews}\\long-form summaries, varied sources};
    \node[proc, right=of over] (parse) {\textbf{Parse attributions}\\extract attribution-citation units};
    \node[proc, right=of parse] (edit) {\textbf{Adversarial editing}\\$\sim$60\% of claims, 19 strategies};
    \node[data, below=1.1cm of edit] (bench) {\textbf{Deep-Research Citation Benchmark}\\624 attribution-citation units};
    \node[proc, left=of bench] (council) {\textbf{Council of 6 LLM judges}\\relevance and factual support};
    \node[hstep, left=of council] (hr) {\textbf{Human review of every decision}\\all 1{,}248 decisions validated};
    \node[leaf, below=1.0cm of hr, xshift=-1.7cm, fill=orange!8] (unan) {Unanimous (870)\\confirmed on review};
    \node[leaf, below=1.0cm of hr, xshift=1.7cm, fill=orange!22] (dis) {Disagreements (378)\\human adjudication};
    \node[data, below=2.8cm of hr] (gold) {\texttt{golden\_dataset.json}\\1{,}872 labels, 1{,}248 judged decisions};

    \draw[arr] (topic) -- (over);
    \draw[arr] (over) -- (parse);
    \draw[arr] (parse) -- (edit);
    \draw[arr] (edit) -- (bench);
    \draw[arr] (bench) -- (council);
    \draw[arr] (council) -- (hr);
    \draw[arr] (hr.south) -- (unan.north);
    \draw[arr] (hr.south) -- (dis.north);
    \draw[arr] (unan.south) -- (gold.north);
    \draw[arr] (dis.south) -- (gold.north);
  \end{tikzpicture}%
  }
  \caption{Construction pipeline for the Deep-Research Citation Benchmark. 25 topics are selected and clean overviews
  are generated with attributed sources. About 60\% of the attributed claims are then
  adversarially edited using 19 base strategies, with some claims receiving more than one,
  while the rest are left clean. Gold labels are produced by a council of 6 LLM judges that
  independently run the source attribution pipeline. A human reviewer validates every council
  decision. The 870 unanimous decisions are confirmed on review, and the 378 non-unanimous
  decisions receive intensive adjudication.}
  \label{fig:golden_workflow}
\end{figure}
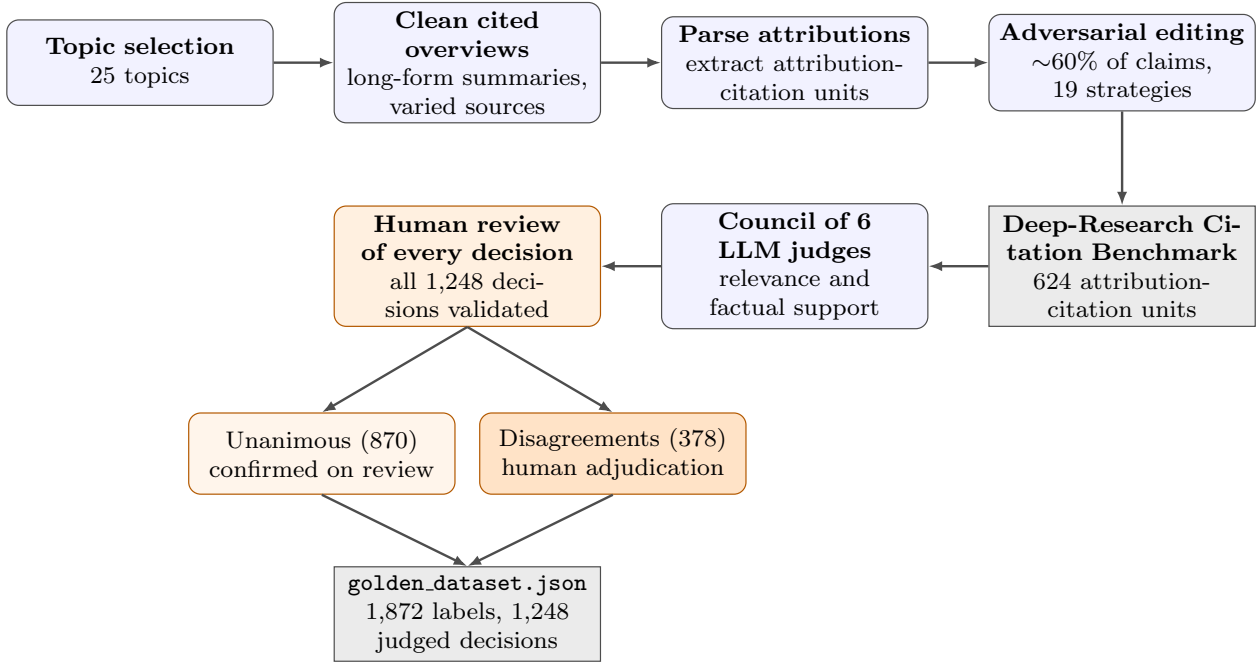

\subsection{Benchmark Overview}
\label{subsec:overview}
The Deep-Research Citation Benchmark is a single long-form report spanning 25 topic domains, namely Amtrak,
Pok\'{e}mon, 1600s historical events, calculus, agentic artificial intelligence, mascots,
neurology, black holes, artisan sourdough baking, quantum computing, the history of the
Silk Road, beekeeping and honey production, urban planning and zoning laws, classical Greek
pottery, electric vehicle battery recycling, Shakespeare's comedies, deep-sea hydrothermal
vents, minimalist interior design principles, traditional Bhutanese music, Antarctic penguin
species, cryptocurrency regulation, flight mechanics of birds, vintage watch collecting,
genetic engineering in agriculture, and the history of the postal system. Domain diversity
reduces the risk that a judge's domain knowledge, rather than its rubric-following
ability, drives performance. Most attributed claims are left clean, while the rest carry intentional factual errors,
live-but-irrelevant sources, and plausible-but-unsupporting sources designed to
stress-test each rubric criterion independently (Figure~\ref{fig:golden_workflow}). Parsing the report with the
pipeline yields 624 attribution-citation pairs.

\begin{table}[t]
\centering
\small
\begin{tabular}{lll}
\toprule
\textbf{Model} & \textbf{Provider} & \textbf{API identifier} \\
\midrule
Claude Haiku~4.5      & Google Vertex & \texttt{claude-haiku-4-5-20251001} \\
Claude Sonnet~4.6     & Google Vertex & \texttt{claude-sonnet-4-6} \\
Claude Opus~4.6       & Google Vertex & \texttt{claude-opus-4-6} \\
Gemini~3.1 Flash Lite & Google Vertex & \texttt{gemini-3.1-flash-lite-preview} \\
Gemini~3.1 Pro        & Google Vertex & \texttt{gemini-3.1-pro-preview} \\
GPT-5-mini            & Azure OpenAI  & \texttt{gpt-5-mini-2025-08-07} \\
GPT-5.4-mini          & OpenAI        & \texttt{gpt-5.4-mini-2026-03-17} \\
GPT-OSS-120B          & AWS Bedrock   & \texttt{gpt-oss-120b} \\
\bottomrule
\end{tabular}
\caption{The 8 LLM judges evaluated, with serving provider and exact API identifier.}
\label{tab:models}
\end{table}

\subsection{Worked Example}
To illustrate the pipeline, consider an adversarial claim, \textit{``The Taylor series
is a finite sum of terms used to approximate functions near a point''} cited to
\url{openstax.org/books/calculus-volume-1/}. Stage~1 extracts this as an
attribution-citation pair, and Stage~2 evaluates each LLM-judged criterion in turn.
\begin{enumerate}
  \item \textit{Source relevance}. The OpenStax calculus page covers Taylor series, so
    this criterion passes.
  \item \textit{Factual support}. The source defines a Taylor series as an
    \textit{infinite} sum of terms, contradicting the claim of a finite sum, so this
    criterion fails.
\end{enumerate}
The structured output is \texttt{\{relevant\_content: 1, fact\_check: 0\}}, with the judge's
rationale noting the mismatch between finite and infinite.

\subsection{Ground-Truth Labels}
Link accessibility labels are produced deterministically by an HTTP check that passes when
the cited URL returns a 200 status, so this criterion needs no LLM and no human review, and
we exclude it from the judge comparison. Source relevance and factual support labels were
produced by a council of 6 LLM judges that independently ran the evaluation pipeline. The
council comprised GPT-5-mini (\texttt{gpt-5-mini-2025-08-07}), GPT-5.2
(\texttt{gpt-5.2-2025-12-11}), Claude Opus~4.5 (\texttt{claude-opus-4-5}), Claude Opus~4.6
(\texttt{claude-opus-4-6}), Gemini~2.5 Pro (\texttt{gemini-2.5-pro}), and Gemini~3 Pro
(\texttt{gemini-3-pro}). A human reviewer then examined the full set of council decisions,
corrected errors caught on review, and adjudicated the final label for every case in which
the council was not unanimous. Of the 378 non-unanimous decisions, 263 were source relevance
and 115 were factual support, with none on link accessibility. Relevance disagreements most
often arose when a source was topically correct but factually wrong, a conflation the
relevance criterion does not explicitly separate from factual support. Agreement among the
labeling judges, the structure of the disagreements, and the resulting gold pass rates are
reported with the other quantitative results in Section~\ref{sec:results}.


\section{Experiments and Results}
\label{sec:results}

\subsection{Experimental Setup}
We evaluate 8 LLM judges from 3 model families (Anthropic, Google, and OpenAI), accessed
through their respective cloud inference APIs. Table~\ref{tab:models} lists each judge with
its serving provider and exact API identifier. Each judge is applied to all 624 source
relevance pairs and all 624 factual support pairs in the benchmark, yielding 1,248 decisions.
We measure pass-class F1 and Cohen's $\kappa$ per dimension as the primary accuracy metrics,
and to characterize reward signal quality beyond accuracy we also measure pass-rate drift,
false positive rate (FPR), and false negative rate (FNR), the bias metrics that determine
what an RLVR training loop over-rewards and under-rewards. All reported F1 and $\kappa$ values
are computed over the 624 pairs per dimension. Two of the 8 judges (GPT-5-mini and Claude
Opus~4.6) also contributed to the gold-label council, but because they span the cost range
and the human-adjudicated subset offers a council-independent comparison, this overlap does
not drive the cost and quality findings.

\begin{table}[t]
\centering
\small
\begin{tabular}{lrcccc}
\toprule
\multirow{2}{*}{\textbf{Model}} &
  \multirow{2}{*}{\textbf{Cost (log$_{10}$, rel.)}} &
  \multicolumn{2}{c}{\textbf{Source Relevance}} &
  \multicolumn{2}{c}{\textbf{Factual Support}} \\
 & & F1 & $\kappa$ & F1 & $\kappa$ \\
\midrule
GPT-5-mini        & 0.33 & \textbf{0.908} {\scriptsize[.89,.93]} & \textbf{0.636} & 0.710 {\scriptsize[.64,.78]} & 0.649 \\
Claude Opus~4.6   & 1.69 & 0.866 {\scriptsize[.84,.89]} & 0.551 & \textbf{0.750} {\scriptsize[.68,.82]} & \textbf{0.701} \\
GPT-5.4-mini      & 0.74 & 0.888 {\scriptsize[.87,.91]} & 0.542 & 0.671 {\scriptsize[.60,.73]} & 0.580 \\
GPT-OSS-120B      & 0.00 & 0.851 {\scriptsize[.83,.88]} & 0.523 & 0.649 {\scriptsize[.56,.72]} & 0.590 \\
Gemini~3.1 Pro    & 1.15 & 0.819 {\scriptsize[.79,.85]} & 0.463 & 0.670 {\scriptsize[.59,.75]} & 0.615 \\
Gemini~3.1 Flash Lite & 0.25 & 0.823 {\scriptsize[.79,.85]} & 0.461 & 0.724 {\scriptsize[.66,.78]} & 0.653 \\
Claude Haiku~4.5  & 0.97 & 0.746 {\scriptsize[.71,.78]} & 0.343 & 0.708 {\scriptsize[.64,.77]} & 0.632 \\
Claude Sonnet~4.6 & 1.48 & 0.700 {\scriptsize[.66,.74]} & 0.322 & 0.726 {\scriptsize[.65,.79]} & 0.675 \\
\bottomrule
\end{tabular}
\caption{Pass-class F1 (with 95\% bootstrap confidence interval, 2{,}000 resamples over
the 624 pairs) and Cohen's $\kappa$ for 8 LLM judges on source relevance and
factual support, with a relative log-cost index per decision. The cost column is a base-10
log index relative to the lowest-cost included model (GPT-OSS-120B, index 0.00), so
$+1.00$ indicates 10$\times$ higher cost. Cost estimates are derived from each provider's
published API pricing as of June 2026, assuming one call per judged dimension per
attribution-citation pair. On factual support every interval overlaps, so no model is
statistically distinguishable on that dimension. Link accessibility is deterministic and
excluded. Best per column in bold.}
\label{tab:main}
\end{table}

Both judge scores and gold labels are binarized at a threshold of $0.5$, where a score
$\geq 0.5$ is treated as \emph{pass}, the positive class. For each dimension we form the
confusion counts TP, FP, FN, TN against the gold label and report the following metrics.
\begin{itemize}
  \item \textbf{Pass-class F1} $= \dfrac{2\,\mathrm{TP}}{2\,\mathrm{TP} + \mathrm{FP} + \mathrm{FN}}$, the harmonic mean of pass-class precision $\mathrm{TP}/(\mathrm{TP}+\mathrm{FP})$ and recall $\mathrm{TP}/(\mathrm{TP}+\mathrm{FN})$.
  \item \textbf{Cohen's} $\kappa = \dfrac{p_o - p_e}{1 - p_e}$, where $p_o$ is the observed judge-gold agreement and $p_e$ is the agreement expected by chance from the marginal pass and fail rates.
  \item \textbf{False positive rate} $\mathrm{FPR} = \dfrac{\mathrm{FP}}{\mathrm{FP} + \mathrm{TN}}$ (a bad citation accepted) and false negative rate $\mathrm{FNR} = \dfrac{\mathrm{FN}}{\mathrm{FN} + \mathrm{TP}}$ (a good citation rejected).
  \item \textbf{Pass-rate drift} $\Delta = r_{\text{judge}} - r_{\text{gold}}$, the signed difference between the judge and the gold pass rate, indicating systematic leniency ($\Delta > 0$) or strictness ($\Delta < 0$).
\end{itemize}
The cost and quality figure uses a \emph{class-balanced average F1}. For each dimension we
average the pass-class and fail-class F1 (macro-F1), then average across the two dimensions,
and we plot this against the estimated cost per judge decision on a logarithmic axis.
Confidence intervals are the $2.5$th and $97.5$th percentiles of the metric over 2{,}000
bootstrap resamples of the pairs.

\subsection{Main Results}

The gold labels reflect the adversarial design of the benchmark. Link accessibility passes
at 98.4\%, source relevance at 79.3\%, and factual support at only 18.4\%, the last kept
intentionally low to stress-test judge discrimination on the hardest citation quality
dimension.

\begin{figure}[t]
  \centering
  \includegraphics[width=13cm]{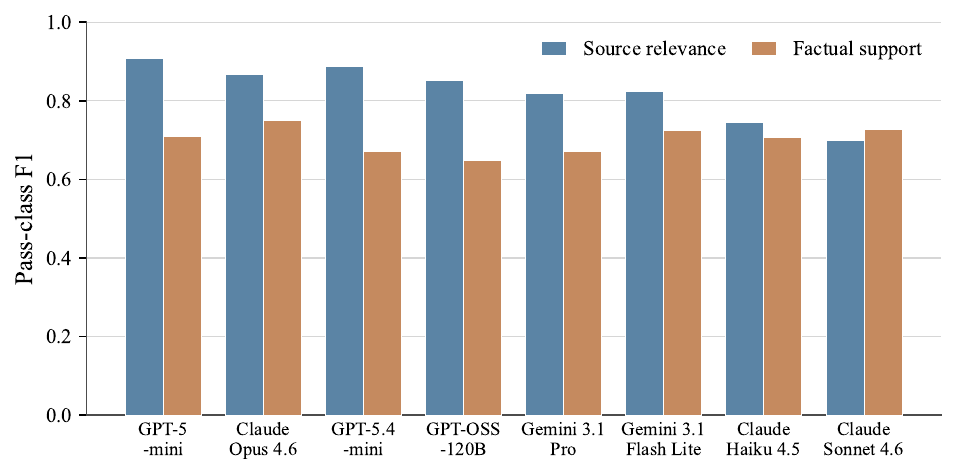}
  \caption{Pass-class F1 for 8 judges on source relevance and factual support.
  No single judge dominates both dimensions.}
  \label{fig:f1}
\end{figure}

Table~\ref{tab:main} reports F1, Cohen's $\kappa$, and the relative log-cost index for all 8
judges, allowing cost and performance to be read per dimension. On source
relevance, performance is heterogeneous, with F1 varying from 0.700 (Claude Sonnet~4.6) to
0.908 (GPT-5-mini) across the evaluated judges.
On factual support, the hardest dimension with a gold pass rate of only 18.4\%,
Claude Opus~4.6 posts the highest point estimate (F1=0.750, $\kappa$=0.701) and
GPT-OSS-120B the lowest (F1=0.649), but every 95\% confidence interval
overlaps on this dimension (from 0.649 [.56,.72] to 0.750 [.68,.82]), so no model is
statistically distinguishable and the apparent Opus lead is within noise. Source
relevance is better resolved, with GPT-5-mini at 0.908 [.89,.93] separating from
lower-ranked models such as Claude Sonnet~4.6 at 0.700 [.66,.74]. GPT-5-mini leads on source relevance (F1=0.908) but ranks fourth on factual support
(F1=0.710), while Claude Opus~4.6 leads on factual support (F1=0.750) but ranks third
on source relevance (F1=0.866), so no single model dominates across both dimensions,
as Figure~\ref{fig:f1} visualizes. Treating judge selection as a single
model choice, rather than a per-criterion decision, leaves reward signal quality on the
table.

\begin{figure}[t]
  \centering
  \includegraphics[width=14cm]{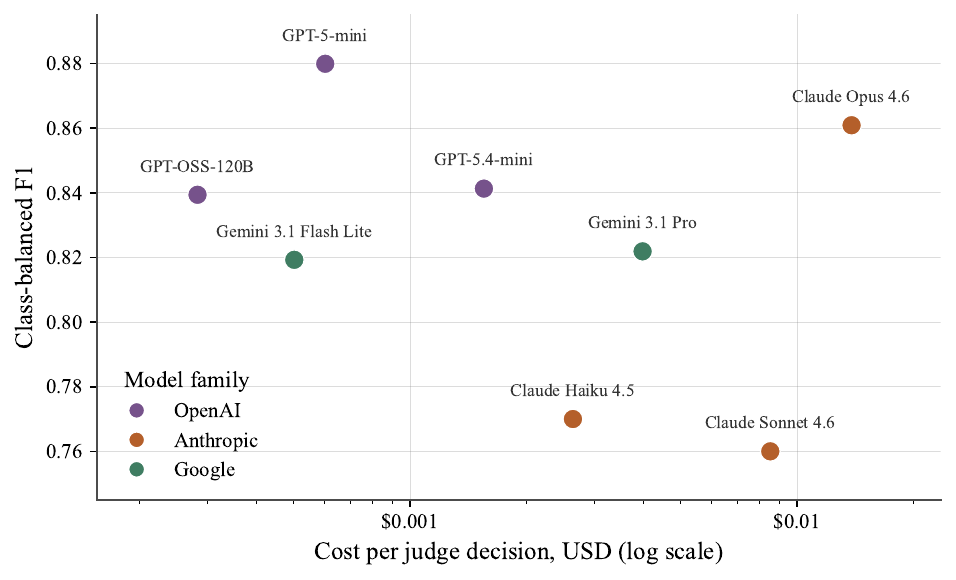}
  \caption{Judge quality (average class-balanced F1), estimated cost per decision
  on a logarithmic axis, and pass-rate drift. Cost does not track quality. Low-cost
  judges are among the most competitive, and cost alone does not predict reward signal
  quality.}
  \label{fig:cost_quality}
\end{figure}

\subsection{Multi-Judge Disagreement Cases}

The 378 multi-judge disagreement cases are the hardest instances in the benchmark, the pairs
where the 6 gold-label judges did not agree and a human adjudicated the final label. Treating
this adjudicated subset as an ablation isolates judge behavior on genuinely ambiguous
citations. Most disagreements were near-consensus rather than deep splits, with 44\% showing
a single dissenting judge and only 12\% evenly split, while 29\% included at least one partial
(0.5) score, indicating that the binary rubric is occasionally applied as a graded one.
Agreement among the labeling judges was substantial, with mean pairwise Cohen's $\kappa$ of
0.62 (range 0.50 to 0.72) on source relevance and 0.67 (range 0.54 to 0.83) on factual
support. On the adjudicated cases, no single labeling judge reliably matched the human label,
with per-model agreement ranging from 38\% to 85\% on source relevance and 43\% to 67\% on
factual support.

\begin{figure}[t]
  \centering
  \includegraphics[width=13cm]{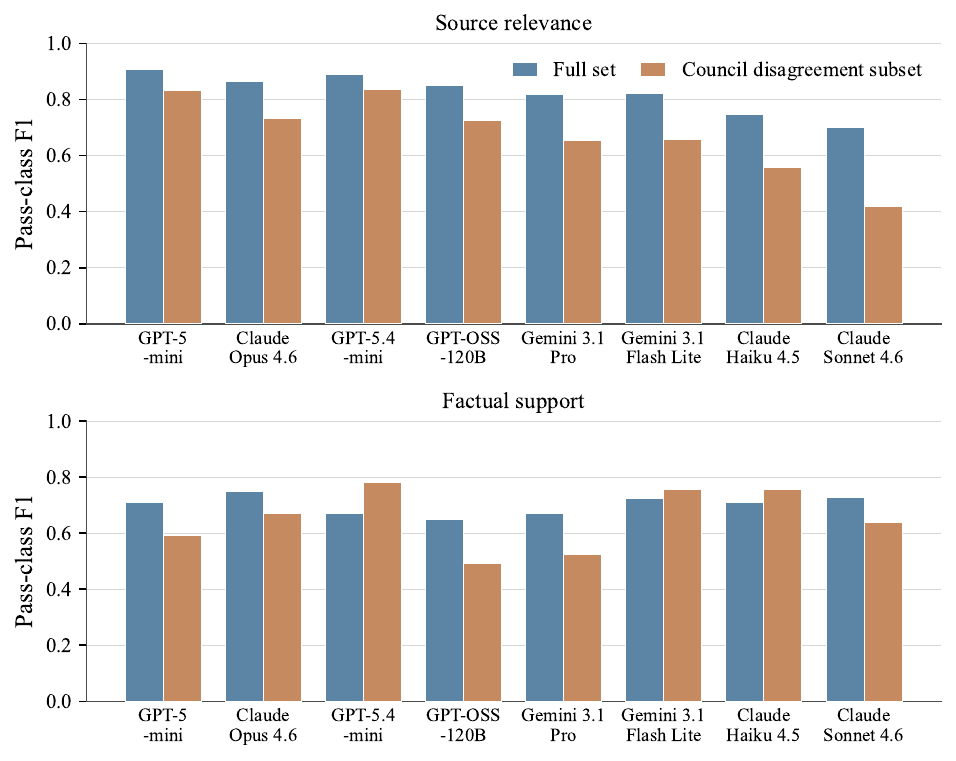}
  \caption{Judge accuracy on the full benchmark compared with the 378 multi-judge
  disagreement cases. Rankings shift substantially on the adjudicated subset, so full-set F1
  is an incomplete proxy for reliability on the most ambiguous citations.}
  \label{fig:hr_degradation}
\end{figure}

On this subset, all 8 judges degrade on source relevance, while factual support
behavior is mixed, with GPT-5.4-mini, Claude Haiku~4.5, and Gemini~3.1 Flash Lite
improving on the adjudicated cases. On source relevance, GPT-5-mini remains the strongest
(F1=0.832) but drops 0.076 from its full-set score, while Claude Sonnet~4.6 degrades most
severely, falling from F1=0.700 to 0.420. On factual support, GPT-5.4-mini leads the subset
(F1=0.780), improving from its full-set score of 0.671 and displacing Claude Opus~4.6
(F1=0.672), which led the full set. Rankings shift substantially between the full set and
this hard subset (Figure~\ref{fig:hr_degradation}). Among the 8 evaluated judges, pairwise
agreement ranges from 76.9\% (GPT-5.4-mini and Claude Sonnet~4.6, $\kappa$=0.535) to 93.6\%
(GPT-OSS-120B and Gemini~3.1 Pro, $\kappa$=0.860), with 21 of 28 pairs between 82\% and 88\%.
The lowest agreement involves the pair with the largest pass-rate difference, where
GPT-5.4-mini passes 49.2\% of evaluated pairs while Sonnet~4.6 passes 28.4\%.

\section{Discussion}
\label{sec:discussion}

\subsection{Directional Bias}

\begin{figure}[t]
  \centering
  \includegraphics[width=13cm]{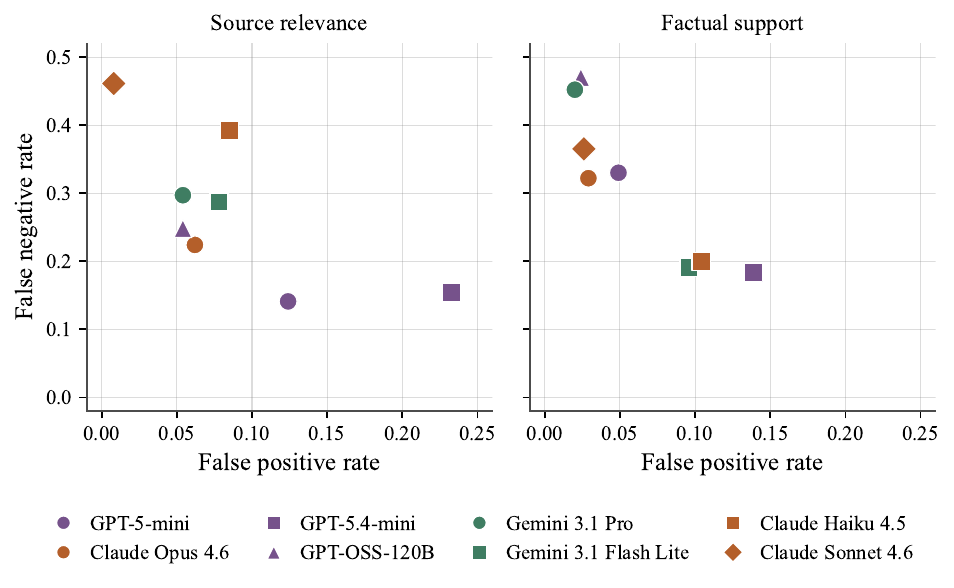}
  \caption{False positive rate (FPR) and false negative rate (FNR) per judge per
  dimension. Judges with similar F1 can differ substantially in directional bias,
  a property invisible to scalar F1 but directly relevant to what an RLVR
  training loop reinforces.}
  \label{fig:fpr_fnr}
\end{figure}

Most judges are stricter than the gold labels on source relevance. All 8
models produce predicted pass rates below the gold pass rate of 79.3\% (from 42.9\% to
72.0\%), indicating a systematic tendency to under-reward citations on that dimension.
Factual support shows more variation, with 3 models (GPT-5.4-mini, Claude Haiku~4.5,
and Gemini~3.1 Flash Lite) exceeding the
18.4\% gold pass rate, reflecting heterogeneous bias across dimensions.
Pass-rate drift is consequential for RLVR training, since a judge that consistently
under-rewards will produce signal sparsity, potentially training models to over-hedge
or under-cite. On source relevance, judges exhibit high precision but
moderate recall, rarely marking irrelevant sources as relevant while frequently missing
relevant ones. On factual support, false negative rates vary from 0.183 (GPT-5.4-mini)
to 0.470 (GPT-OSS-120B), indicating that most judges reject a substantial
fraction of genuinely supported citations. As shown in Figure~\ref{fig:fpr_fnr}, FNR
profiles differ substantially across judges even at similar F1 scores, confirming
that scalar F1 obscures the reward signal asymmetries most consequential for training
loop design. This asymmetry maps directly onto the reward hacking mechanism studied in
rubric RL~\citep{mahmoud2025,primeintellect2026}, where a judge with high FPR over-rewards bad citations,
driving the trained model to exploit the judge's permissive behavior rather
than learn the true quality signal. A judge with high FNR instead under-rewards genuinely
supported claims, producing signal sparsity and potentially training models to
over-hedge or under-cite.

\subsection{Difficulty by Adversarial Strategy}

To locate the factual-support difficulty, we linked each adversarially edited claim to its
strategy (Appendix~\ref{app:strategies}) and measured how often the 8 judges reject it.
Detection is uniformly high, from roughly 86\% for the subtlest edits (wrong attribution,
minimization, and partial truth) to near 100\% for negations and semantic drift, so judges
rarely accept an edited citation. The low pass-class factual-support F1 instead reflects the
opposite error, over-rejection of genuinely supported citations, consistent with the high
false negative rates above. This distinction matters for reward design, because the dominant
factual-support failure is under-rewarding correct citations rather than over-rewarding
fabricated ones, so calibration effort should target judge strictness rather than
fabrication detection.

\subsection{Cost and Quality}

Judge cost per decision spans a 49$\times$ range across the 8 models,
from GPT-OSS-120B to Claude Opus~4.6, and cost does not predict accuracy.
GPT-5-mini, the third-cheapest model, achieves the highest source relevance F1
(0.908), making it the most accurate on one of the two key dimensions.
Gemini~3.1 Flash Lite, the second-cheapest, achieves an average class-balanced F1 of 0.771,
competitive with models up to 8$\times$ more expensive. As shown in
Figure~\ref{fig:cost_quality}, cost does not track quality. Mid-tier models are not
consistently competitive with either the cheapest or the most accurate options, so judge
selection should be driven by per-dimension rubric fit rather than model tier or price.
Cost estimates are derived from each provider's published API pricing as of June 2026,
assuming one LLM call per judged dimension per attribution-citation pair.
Batch scoring, which evaluates all criteria in a single call, and prompt
caching~\citep{lumer2026don}, which amortizes shared context across tool calls in
long-horizon agentic workloads, would reduce costs further~\citep{harvey2026}. More
broadly, judge selection is a component of the agent design problem in production systems,
where the choice of model, tool, and scoring strategy must be made jointly~\citep{lumer2026tool}.

\subsection{Reward Noise from Disagreement}

The multi-judge disagreement subset (Section~\ref{sec:results}) shows that full-set F1 is an
incomplete proxy for judge reliability on the ambiguous cases most likely to determine reward
signal quality in practice, since rankings shift there and no single model reliably matches
the adjudicated label, so single-model consensus is insufficient and human adjudication is
essential. Low inter-judge agreement signals high reward noise, because decisions on which
judges disagree introduce inconsistent training signal for the same input. Selecting judges
with high pairwise agreement, or ensembling across disagreeing judges, can reduce this noise
without a stronger model.


\section{Conclusion}
\label{sec:conclusion}

As reinforcement learning increasingly delegates reward modeling to LLM judges, a practical
question follows about how capable the judge needs to be and how much its bias matters.
We study this calibration question for citation quality in deep-research systems, where each
attributed claim must be judged along two LLM-scored dimensions, source relevance and factual
support, using the Deep-Research Citation Benchmark of 1,248 human-reviewed rubric decisions.
Across 8 off-the-shelf judges from 3 model families, cheaper judges prove competitive on
source relevance, with GPT-5-mini (F1=0.908, 95\% CI [.89,.93]) leading among the
lowest-cost options, while on factual support all confidence intervals overlap and no judge
is statistically distinguishable from the rest. Beyond accuracy, judges differ substantially
in pass-rate drift, false positive rate, and false negative rate at similar F1 levels, and
these directional biases shape what a downstream RL loop would reinforce in ways that scalar
F1 does not reveal. These findings are limited to a single adversarial document, and
sensitivity to prompt design and batching remain open questions. Calibrating the judge is a
prerequisite for using citation rubrics as reward signals, and our results show that this
calibration does not require the most expensive model available.


\FloatBarrier
\bibliography{references}

\newpage
\appendix

\section{Judge Prompts and Rubric}
\label{app:prompts}

Every LLM-judged criterion is scored by a single LLM call with the system and
human prompts below. Link accessibility is not prompted, since it is a deterministic HTTP
check that passes when the cited URL returns a 200 status. Both prompts request a binary
score (1 = criterion met, 0 = not met) together with a free-text rationale, and are applied
identically across all 8 judges.

\paragraph{Source relevance.}
\begin{verbatim}
SYSTEM:
You are an expert evaluator assessing content relevance.
Your task is to determine if the content at a URL is relevant to a given
attribution text.

Evaluate whether the content at the URL is relevant to the attribution text.
Consider:
- Does the URL content discuss the same topic as the attribution?
- Does it provide supporting information for the claims in the attribution?
- Is there a clear connection between the attribution and the URL content?

This is a binary evaluation:
- score = 1 if the content IS relevant
- score = 0 if the content is NOT relevant

HUMAN:
Attribution Text:
{attribution_text}

URL: {url}

URL Content:
{url_content}

Evaluate whether this URL content is relevant to the attribution text.
\end{verbatim}

\paragraph{Factual support.}
\begin{verbatim}
SYSTEM:
You are an expert fact-checker evaluating factual accuracy.
Your task is to determine if the factual claims in an attribution text are
supported by the content at a URL.

Evaluate whether the factual claims in the attribution text are supported by
the URL content. Consider:
- Are the specific facts, numbers, dates, and claims in the attribution text
  present in the URL content?
- Do they match or are they consistent?
- Are there any contradictions?

This is a binary evaluation:
- score = 1 if the facts ARE supported by the URL content
- score = 0 if the facts are NOT supported (contradicted, missing, or
  unverifiable)

HUMAN:
Attribution Text:
{attribution_text}

URL: {url}

URL Content:
{url_content}

Verify whether the factual claims in the attribution text are supported by
this URL content.
\end{verbatim}

The relevance rubric asks whether the source ``provide[s] supporting information for the
claims,'' which overlaps with the factual support criterion. This overlap is the most
frequent source of inter-judge disagreement on relevance (Section~\ref{sec:method}), where
judges split on sources that are topically correct but factually wrong. Tightening the
separation between the two criteria is a concrete direction for improving rubric
specificity.

\section{Adversarial Edit Strategies}
\label{app:strategies}

The Deep-Research Citation Benchmark applies adversarial edits to roughly 60\% of attributed
claims using the 19 strategies below (some claims receive more than one). Clean claims are
left unchanged from the source overview.

\begin{description}\itemsep0pt
  \item[\textsc{wrong\_date}] incorrect dates or years.
  \item[\textsc{wrong\_number}] incorrect statistics, quantities, or distances.
  \item[\textsc{wrong\_entity}] wrong person, company, or organization.
  \item[\textsc{negation}] claim asserts the opposite of the truth.
  \item[\textsc{wrong\_location}] incorrect geographic location.
  \item[\textsc{fabricated\_detail}] plausible but invented specifics.
  \item[\textsc{wrong\_attribution}] credited to the wrong person or organization.
  \item[\textsc{magnitude\_error}] off by orders of magnitude.
  \item[\textsc{source\_mismatch}] citation URL points to a real but wrong-topic page.
  \item[\textsc{irrelevant\_claim}] a fully off-topic claim is inserted.
  \item[\textsc{semantic\_drift}] meaning is subtly shifted from the truth.
  \item[\textsc{reversed\_causation}] cause and effect are swapped.
  \item[\textsc{near\_miss}] almost correct but one key detail is wrong.
  \item[\textsc{cross\_topic}] a fact from a different topic is inserted.
  \item[\textsc{partial\_truth}] true and false elements are mixed.
  \item[\textsc{overgeneralization}] a specific truth is extended too broadly.
  \item[\textsc{outdated\_info}] old information is presented as current.
  \item[\textsc{exaggeration}] numbers or impact are inflated.
  \item[\textsc{minimization}] numbers or impact are understated.
\end{description}

\end{document}